\documentclass[10pt,twocolumn,letterpaper]{article}
\usepackage[accsupp]{axessibility}

\usepackage{iccv}
\usepackage{times}
\usepackage{epsfig}
\usepackage{graphicx}
\usepackage{amsmath}
\usepackage{amssymb}

\usepackage{caption}
\usepackage{booktabs}
\usepackage{multirow,array}
\usepackage{algorithm}
\usepackage{algorithmic}

\usepackage{nccmath}
\usepackage{tabularx}
\usepackage{xltabular}
\usepackage{booktabs}
\usepackage{color}

\newcommand\blfootnote[1]{%
  \begingroup
  \renewcommand\thefootnote{}\footnote{#1}%
  \addtocounter{footnote}{-1}%
  \endgroup
}

\iccvfinalcopy 

\ificcvfinal\fi

\begin{document}

\title{Manifold Alignment for Semantically Aligned Style Transfer}


\author{Jing Huo$^1$, Shiyin Jin$^1$, Wenbin Li$^{1*}$, Jing Wu$^2$, Yu-Kun Lai$^2$, Yinghuan Shi$^1$, Yang Gao$^1$\\
$^1$State Key Laboratory for Novel Software Technology, Nanjing University\\
$^2$School of Computer Science \& Informatics, Cardiff University\\
}

\twocolumn[{%
\renewcommand\twocolumn[1][]{#1}%
\maketitle
\begin{center}
    \centering
    \includegraphics[width=0.98\textwidth]{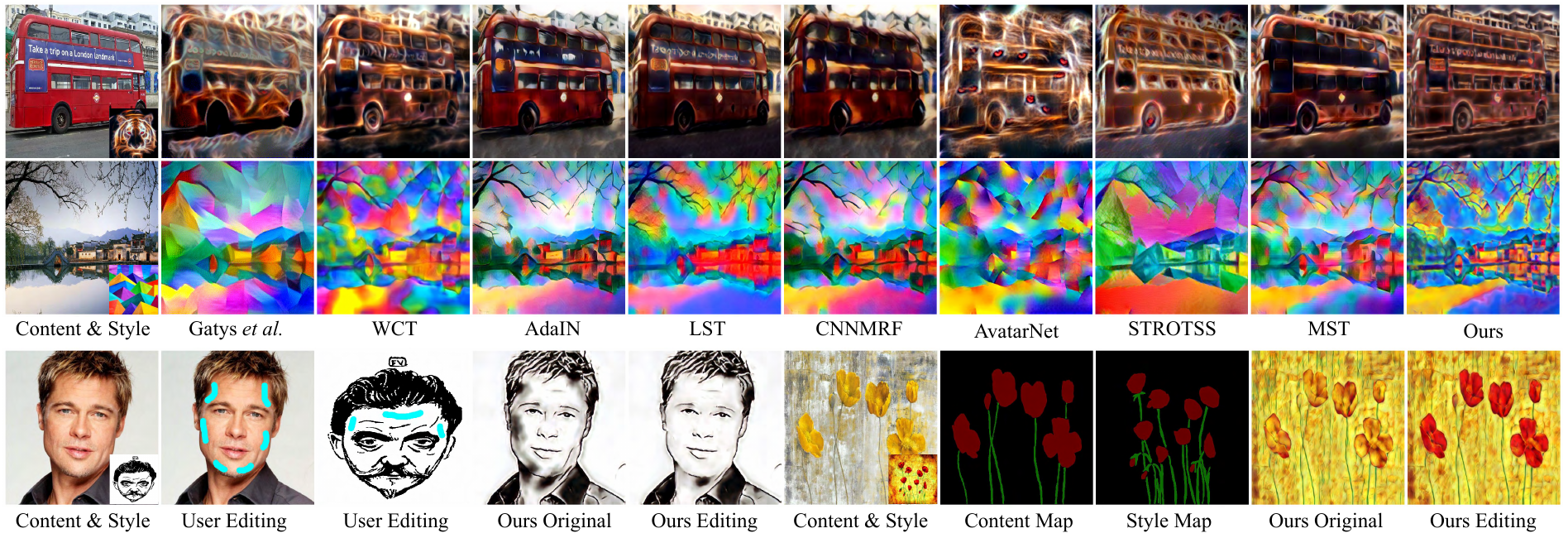}
    \setlength{\abovecaptionskip}{0.01cm}
    \captionof{figure}{\small{The first two rows show comparisons with some state-of-the-art (SOTA) methods. Three kinds of SOTA methods are compared, where Gatys \etal~\cite{gatys2016image}, WCT~\cite{li2017universal}, AdaIN~\cite{huang2017arbitrary}, LST~\cite{li2019learning} are global statistics based methods, CNNMRF~\cite{li2016combining} and AvatarNet~\cite{sheng2018avatar} are local patch based methods, and STROTSS~\cite{kolkin2019style} and MST~\cite{Zhang_2019_ICCV} are semantic region based methods. The proposed method is the best in semantic structure preservation. The third row shows our user editing and semantic map guided style transfer results.}}
    \label{Fig_FirstFigure}
\end{center}%
}]


\ificcvfinal\thispagestyle{empty}\fi

\blfootnote{$^*$Corresponding author.}

\begin{abstract}
Most existing style transfer methods follow the assumption that styles can be represented with global statistics (\emph{e.g.,} Gram matrices or covariance matrices), and thus address the problem by forcing the output and style images to have similar global statistics. An alternative is the assumption of local style patterns, where algorithms are designed to swap similar local features of content and style images. However, the limitation of these existing methods is that they neglect the semantic structure of the content image which may lead to corrupted content structure in the output. In this paper, we make a new assumption that image features from the same semantic region form a manifold and an image with multiple semantic regions follows a multi-manifold distribution. Based on this assumption, the style transfer problem is formulated as aligning two multi-manifold distributions and a \textit{Manifold Alignment based Style Transfer (MAST)} framework is proposed. The proposed framework allows semantically similar regions between the output and the style image share similar style patterns. Moreover, the proposed manifold alignment method is flexible to allow user editing or using semantic segmentation maps as guidance for style transfer. To allow the method to be applicable to photorealistic style transfer, we propose a new adaptive weight skip connection network structure to preserve the content details. Extensive experiments verify the effectiveness of the proposed framework for both artistic and photorealistic style transfer. Code is available at https://github.com/NJUHuoJing/MAST. 
\end{abstract}

\section{Introduction}
The goal of style transfer is to synthesize an output image by transferring the target style to a given content image. Currently, most methods~\cite{gatys2016image, huang2017arbitrary,  li2017universal, li2019learning} make the assumption that image styles can be represented by global statistics of deep features, such as Gram matrices or covariance matrices. Such global statistics capture the style from the whole image, and are applied to the content image without differentiation of the contents inside. However, for images containing different semantic parts, such global statistics are insufficient to represent the multiple styles required for proper style transfer. As shown in the first two rows of Figure~\ref{Fig_FirstFigure}, although the overall appearances of the results from Gatys \etal~\cite{gatys2016image}, AdaIN~\cite{huang2017arbitrary}, WCT~\cite{li2017universal} and LST~\cite{li2019learning} look like the style image, they sometimes fail to preserve the local semantic structure of the content image, leading to corrupted image content. Another kind of style transfer methods is the local patch based methods~\cite{chen2016fast, sheng2018avatar, li2016combining, liao2017visual, gu2018arbitrary}, where the assumption is that local features of the content image can be replaced with local features of the style image to produce stylized output. However, the local content features may be mismatched to features that do not share similar semantic meanings, leading to artifacts~\cite{Zhang_2019_ICCV}. See the result of AvatarNet~\cite{sheng2018avatar} in the first row of Figure \ref{Fig_FirstFigure}, where the tiger's eyes appear on the bus in the stylized image. Therefore, in many situations, neither the global statistics based nor local patch based methods are suitable. Recently, some works~\cite{Zhang_2019_ICCV,kolkin2019style} have proposed style transfer based on locally aligned semantics and have achieved better results in terms of content structure preserving and style transfer within similar semantic regions. However, existing works either contain multiple stages~\cite{Zhang_2019_ICCV} or have many terms to balance \cite{kolkin2019style}, making the overall algorithm inefficient and hard to tune. To address these limitations, we make a new assumption in transferring different styles and propose a simple and efficient style transfer method with impressive results. 


Specifically, in this paper, we make the assumption that image features from the same semantic region form a single manifold. Therefore, for an image with multiple objects, all the features in this image follow a multi-manifold distribution. The style transfer problem thus becomes a problem of aligning two multi-manifold distributions of the style and the content features. Based on this assumption, we propose a \textit{manifold alignment based style transfer (MAST)} 
framework. The proposed manifold alignment method is based on subspace learning, and learns a projection matrix to project the content features into the style features' subspace. In the style features' subspace, content features and style features sharing the same or similar semantic meanings (\textit{i.e.}, having large feature similarities) are forced to be close, \textit{i.e.}, the locally aligned semantic information between content and style features is preserved. This makes the semantically aligned image regions in the output and the style images have similar style patterns. The proposed manifold alignment method can be easily plugged into many existing auto-encoder based style transfer structures~\cite{li2017universal}. Example results are given in the 1st and 2nd rows of Figure~\ref{Fig_FirstFigure}. Moreover, by using user-defined region correspondence of content and style images, the proposed method can be easily extended to support user editing or use semantic segmentation maps as guidance. Examples are given in the 3rd row of Figure~\ref{Fig_FirstFigure}.

Applying the proposed framework to photorealistic style transfer requires fully preserving the details of content images. However, existing auto-encoder based style transfer structures can lead to loss of detailed information and result in distortions in the output. We therefore propose a new adaptive weight skip connection (AWSC) structure to preserve the detailed spatial features of the content image. We further extend our manifold alignment method as an orthogonal constrained optimization problem, which compared with the non-orthogonal version is proved to be able to fully preserve the content structure during style transfer. The new AWSC structure together with the orthogonal constrained manifold alignment is shown effective for photorealistic style transfer. 


The main contributions of this paper are as follows:

Firstly, we introduce the novel view that style transfer can be treated as a manifold alignment problem, and propose a new manifold alignment algorithm for aligning multi-manifold distributions to address the semantically aligned style transfer problem. 

Secondly, we show that the algorithm is flexible to allow user editing or using semantic segmentation maps as guidance in style transfer. Extensive experiments show the proposed algorithm achieves promising results.

Lastly, we extend the algorithm for photorealistic style transfer with a new adaptive weight skip connection structure and orthogonal constraints, which is shown to produce high quality photorealistic style transfer results.

\section{Related Work}
\subsection{Artistic Style Transfer}
According to the style assumption adopted, existing style transfer works~\cite{johnson2016perceptual,wang2017multimodal, luan2017deep,Lu_2019_ICCV} can mainly be categorized into global statistics based~\cite{li2017universal,huang2017arbitrary,li2019learning,Chiu_2019_ICCV}, local patch based~\cite{chen2016fast,li2016combining,sheng2018avatar,gu2018arbitrary} and semantic region based methods~\cite{li2019learning,kolkin2019style}. 



\textbf{Global statistics based style transfer.} 
In early work, style transfer is mainly achieved by forcing the global statistics of the output image to be the same as the style image. Gatys \etal~\cite{gatys2016image} use the Gram matrix to represent the style of an image and use an optimization-based method to achieve the consistency of the Gram matrices of the output image and the style image. Li \etal~\cite{li2017universal} use the
whitening and coloring transforms (WCTs) to align the covariance of the output and the style features. Huang \etal~\cite{huang2017arbitrary} achieve this goal by adopting adaptive instance normalization (AdaIN) to make the mean and variance of the content features the same as those of the style features. Li \etal~\cite{li2019learning} propose to use a feed-forward network to predict the style transformation matrix. However, these methods represent styles as global statistics which, as discussed in the previous section, may not be appropriate in many scenarios.

\textbf{Local patch based style transfer.} Local patch based methods~\cite{liao2017visual,gu2018arbitrary} generally follow the local style pattern assumption and address the problem by swapping local features. Chen \etal~\cite{chen2016fast} propose a method to replace the content feature patches by similar style feature patches. Li and Wand~\cite{li2016combining} propose to incorporate generative Markov Random Field (MRF) models for local patch based style transfer. Sheng \etal~\cite{sheng2018avatar} propose an AvatarNet which combines a whitening operation and a local patch reassembling operation. The major drawback of patch based methods is that undesired artifacts may appear when the local features of content and style images are mismatched.

\textbf{Semantic region aligned style transfer.} Zhang \etal~\cite{Zhang_2019_ICCV} propose to cluster style and content features into different components, which are matched using graph cut. Then WCT is used to achieve style transfer in each matched pair. However, MST cannot be easily extended to allow user editing. Moreover, it contains multiple stages and has many parameters to tune. Kolkin \etal~\cite{kolkin2019style} adopt an optimization based method which balances the optimization between a global style loss and a local distribution aware style loss. It allows user control of the style transfer regions. However, the optimization based framework makes it computationally inefficient. Compared with these two methods, we make a different style transfer assumption and propose a simple and efficient method for style transfer with promising results. 

\subsection{Photorealistic Style Transfer}
Different from artistic style transfer, photorealistic style transfer requires the stylized output to fully preserve the original content details. Luan \etal~\cite{luan2017deep} augment the neural optimization based style transfer method with photorealistic constraints. Li \etal~\cite{li2018closed} extend the WCT for photorealistic style transfer by modifying the decoder structure and introducing a post processing step. Yoo \etal~\cite{yoo2019photorealistic} propose to use wavelet pooling and unpooling to replace the pooling and upsampling operations to preserve the content details. Another work of Li \etal~\cite{li2019learning} uses a linear transformation module for photorealistic style transfer. As can be seen, the focus of photorealistic style transfer is to preserve the content details. We propose a new skip connection based structure to preserve content details which combined with the orthogonal constrained manifold alignment is proved to be effective for photorealistic style transfer. 


\subsection{Manifold Alignment}
The objective of manifold alignment is to align two sets of data from two manifold distributions in a common subspace by leveraging the correspondences between the two, where the correspondence information is usually in the form of pairwise similarity. Depending on whether the correspondence information is provided or not, there are semi-supervised~\cite{DBLP:conf/ijcai/WangM11,DBLP:conf/aistats/HamLS05,lin2006inter} and unsupervised manifold alignment methods~\cite{DBLP:conf/ijcai/WangM09,cui2014generalized,pei2011unsupervised}. Generally, manifold alignment algorithms are designed to learn a common subspace which preserves both the cross manifold correspondence and the original manifold structures. Different from existing methods, the proposed alignment method aligns the two distributions in the style features' subspace instead of the common subspace.

\begin{figure*}[t]
  \centering
  \includegraphics[width=0.85\textwidth]{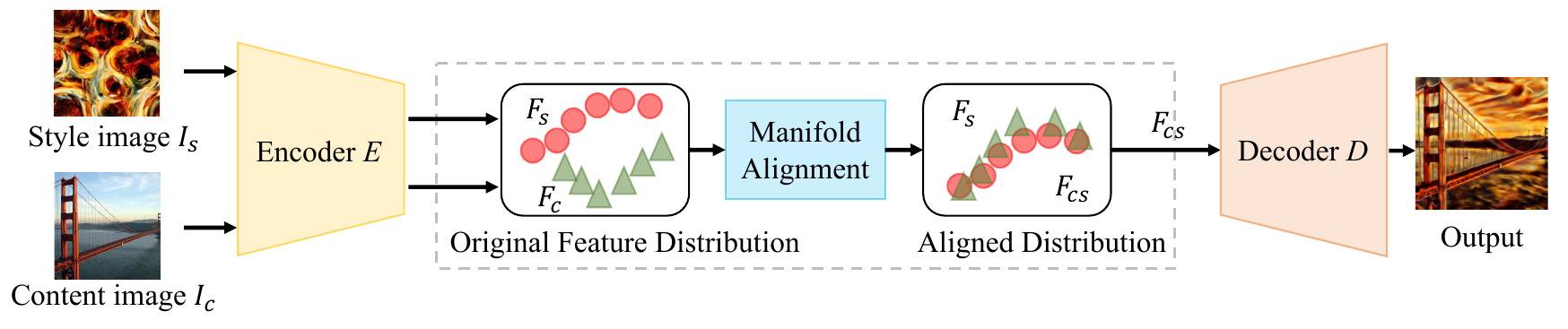}
  \setlength{\abovecaptionskip}{0.01cm}
  \caption{\small{Framework of the proposed manifold alignment based style transfer (MAST). With a pre-trained encoder and decoder, the proposed manifold alignment method uses the output features of the encoder to find a projection such that the content and style features are aligned in the style features' subspace. Then the projected content features are fed into the decoder to reconstruct the stylized output. The proposed manifold alignment module can be plugged into many auto-encoder based style transfer network structures.} }
  \label{Fig_Framework}
\end{figure*}

\section{Manifold Alignment for Style Transfer}
The overall framework of the proposed manifold alignment based style transfer method is given in Figure~\ref{Fig_Framework}. The style transfer is formulated with an encoder, a decoder and feature transformation. In our work, the feature transformation which transforms content features into the style features' subspace is implemented by a manifold alignment algorithm. Section \ref{Sec_MA} gives details of the proposed manifold alignment algorithm. Notice the manifold alignment module can be plugged into many existing auto-encoder based style transfer frameworks. In Section \ref{Sec_STF}, we propose a skip connection based network structure in combination with the manifold alignment method for photorealistic style transfer.  

\subsection{Manifold Alignment Method}
\label{Sec_MA}
Given a content image $I_c$ and a style image $I_s$, define their features extracted by an encoder as $F_c\in \mathbb{R}^{C\times (W_c\times H_c)}$ and $F_s\in \mathbb{R}^{C\times (W_s \times H_s)}$, where $C$ is the number of channels of the feature maps, $W_c$ ($W_s$) and $H_c$ ($H_s$) are the widths and heights of the feature maps. Subscripts $c$ and $s$ refer to the content and style images, respectively. Therefore, $F_c$ ($F_s$) can be seen as a set of $W_c\times H_c$ ($W_s\times H_s$) number of feature vectors, with each feature vector of dimension $C$. Without any processing, the two sets of features are of different multi-manifold distributions, as there may be multiple objects on the images. The objective of our manifold alignment algorithm is to learn a projection matrix $P\in \mathbb{R}^{C\times C}$ to transform the content features into the subspace of style features such that the two distributions are aligned. Denote the transformed features as $ F_{cs} = P^T F_c$, where $F_{cs} \!\in\! \mathbb{R}^{C\times (W_c\times H_c)}$ is the transformed content features in the style features' subspace. 

The objective of the manifold alignment algorithm for finding the projection is that features with the similar semantic meanings in the original subspace are still located closely after the transformation. However, in most cases, the semantic meaning of features is not given. In this case, we adopt the normalized cross correlation~\cite{chen2016fast} between features as the similarity measure to build an affinity matrix. The normalized similarity is defined as $s(x,y)= \frac{x^{T}y}{\|x\|\|y\|}$ with $x\in\mathbb{R}^{C}$, $y\in\mathbb{R}^C$. Specifically, we denote the affinity matrix of the content and the style features as $A^{cs} \in \mathbb{R}^{(W_c\times H_c)\times (W_s\times H_s)}$, with each element of $A^{cs}$ as:
\begin{equation}\label{Eq_A_Cross}\small
  A_{ij}^{cs} = \left\{
\begin{aligned}
1  \quad & \phi_i(F_c) \in \mathcal{N}_k(\phi_j(F_s)) \; \mbox{or} \; \phi_j(F_s) \in \mathcal{N}_k(\phi_i(F_c))\,, \\
0  \quad & \mbox{otherwise}
\end{aligned}
\right. 
\end{equation}
where $\phi_i(F_c)$ denotes the operation to get the $i$th feature from $F_c$ and $\phi_j(F_s)$ is defined similarly. $\mathcal{N}_k(\phi_j(F_s))$ is a set of $\phi_j(F_s)$'s $k$ nearest neighbors in the content feature space. $\mathcal{N}_k(\phi_i(F_c))$ is $\phi_i(F_c)$'s $k$ nearest neighbors in the style feature space. 
We use the normalized similarity between features to find the nearest neighbors. Therefore, $A_{ij}^{cs}$ equals $1$ if $\phi_i(F_c)$ is one of $\phi_j(F_s)$'s $k$-nearest neighbors or vice versa.

\textbf{Objective Functions.} With the above defined affinity matrix, the objective of the proposed manifold alignment method is to make the content feature and style feature with similar or same semantic meanings be close after the content feature is projected into the style feature space. The learning objective function is as follows:
\begin{equation}\small
\begin{aligned}
&\min_{P} J(P) = \frac{1}{N} \sum_{i=1}^{W_c\!\times\! H_c} \sum_{j=1}^{W_s\!\times\! H_s} A_{ij}^{cs} \| \phi_i(F_{cs}) - \phi_j(F_s)\|_2^2,
\end{aligned}
\label{Eq_Obj}
\end{equation}
where $N$ is the number of pairs of nearest neighbors. In the above formulation, when $A_{ij}^{cs}$ equals $1$, $\phi_i(F_{c})$ and $\phi_j(F_s)$ are nearest neighbors in the original space. $\phi_i(F_{cs})$ and $\phi_j(F_s)$ are thus forced to be close in the style feature space.


In addition, to better preserve the content structure of the content image, we can add orthogonal constraints on the projection matrix, with $P^T P = I$, where $I$ is the identity matrix. With $P$ being orthogonal, the similarity of the original content features is always preserved, i.e., $d(\phi_i(F_c), \phi_j(F_c)) = d(\phi_i(F_{cs}), \phi_j(F_{cs}))$. The proof is straight-forward:
\begin{equation}\small
\begin{aligned}
d(\phi_i(F_{cs}), \phi_j(F_{cs})) & =\| P^T \phi_i(F_c) - P^T \phi_j(F_c) \|_2^2\\
& =P^T P \| \phi_i(F_c) - \phi_j(F_c) \|_2^2 \\
&= \| \phi_i(F_c) - \phi_j(F_c) \|_2^2.
\end{aligned}
\end{equation}
According to~\cite{kolkin2019style,shechtman2007matching}, the self-similarities (similarities of features at different locations) of image features encode the image structure. Therefore, the original content structure will always be preserved on the stylized output image.

\begin{figure*}[t]
  \centering
  \includegraphics[width=0.95\textwidth]{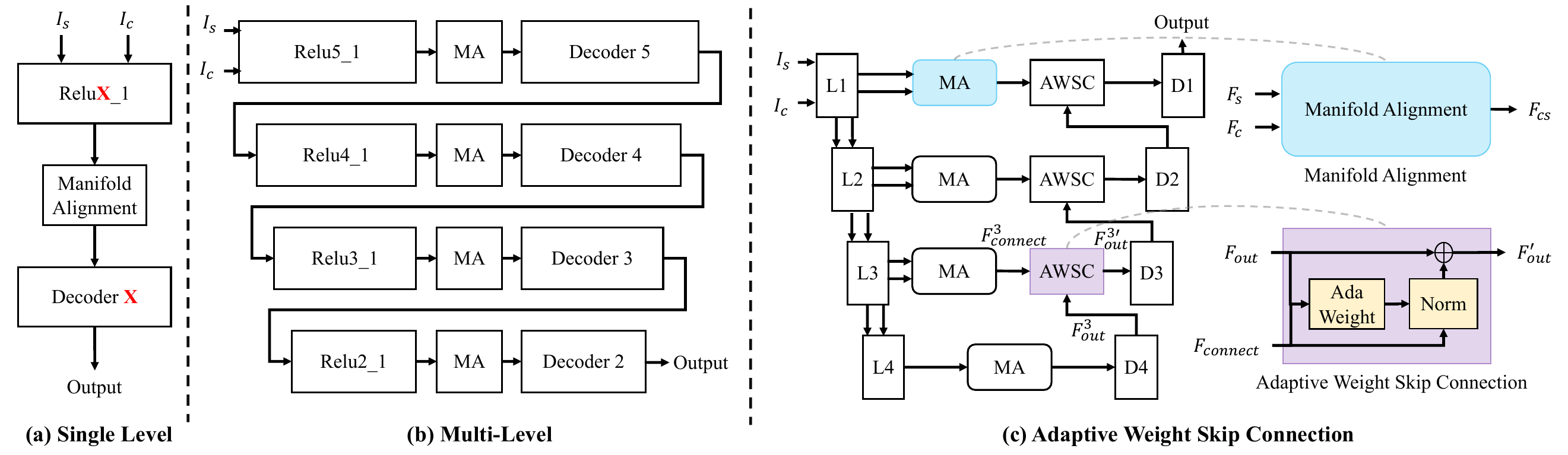}
  \setlength{\abovecaptionskip}{0.01cm}
  \caption{\small{The proposed manifold alignment combined with existing single level style transfer structure (a) and multi-level style transfer structure (b) for artistic style transfer. (c) The proposed adaptive weight skip connection structure for photorealistic style transfer. In (b) and (c), `MA' and `AWSC' denote our manifold alignment and Adaptive Weight Skip Connection modules.} }
  \label{Fig_Framework_GST}
\end{figure*}


\textbf{Optimization of Eq. (\ref{Eq_Obj}).} For both the original and orthogonal constrained optimization problems, we provide efficient closed form solutions as follows. To solve the optimization problem in Eq. (\ref{Eq_Obj}), we can first transform the objective function into the following form. 
\begin{equation}\label{Eq_Obj_Pc}\small
\begin{aligned}
&\min_{P} J(P) =  \mbox{tr}( P^T F_c D_c F_c^T P + F_s D_s F_s^T - 2 P^T F_c U_{cs} F_s^T ),
\end{aligned}
\end{equation}
where $U_{cs} = \frac{1}{N}A^{cs}$, $D_c\!\in\! \mathbb{R}^{(W_c\!\times\! H_c)\times (W_c\!\times\! H_c)}$ is a diagonal matrix, with $D_c(i,i) \!=\! \sum_{j=1}^{(W_s\!\times\! H_s)} U_{cs}(i,j)$. $D_s\!\in\! \mathbb{R}^{(W_s\!\times\! H_s)\times (W_s\!\times\! H_s)}$ is also a diagonal matrix and $D_s(j,j) \!=\! \sum_{i=1}^{(W_c\!\times\! H_c)} U_{cs}(i,j)$. $\mbox{tr}()$ is the trace. In the above equation, the value of the second term is fixed regardless of $P$. Therefore the final objective is to find $P$ that minimizes $J(P) =  \mbox{tr}( P^T F_c D_c F_c^T  - 2 P^T F_c U_{cs} F_s^T )$. To achieve this goal, we take the derivative of $J(P)$ with respect to $P$, 
\begin{equation}\label{Eq_DP}\small
\frac{\partial J}{\partial P} = 2 (F_c D_c F_c^TP - F_c U_{cs} F_s^T),
\end{equation}
By setting the above derivative to 0 and solve the equation, we can derive the closed form solution of $P$ as:
\begin{equation}\label{Eq_SP}\small
P = (F_c D_c F_c^T)^{-1} (F_cU_{cs}F_s^T).
\end{equation}

\textbf{Optimization of Eq. (\ref{Eq_Obj}) with orthogonal constraint.} In this case, $J(P)$ can also be transformed into the formulation in Eq. (\ref{Eq_Obj_Pc}). However, with the orthogonal constraint included, both the first and second terms have values irrelevant to $P$. Therefore, the final objective function is simplified as $ \min_{P} J(P) =  \mbox{tr}( - 2 P^T F_c U_{cs} F_s^T ), \mbox{s.t.}\   P^T P = I$ and is equivalent to:
\begin{equation}\label{Eq_Obj_Pc_Orth}\small
\begin{aligned}
&\max_{P} J(P) =  \mbox{tr}(2 P^T F_c U_{cs} F_s^T )\\
&\mbox{s.t.}\   P^T P = I.
\end{aligned}
\end{equation}
To solve the above optimization problem, we apply singular value decomposition (SVD) to $F_c U_{cs} F_s^T$. Suppose $F_c U_{cs} F_s^T = U\Lambda V^T$, where $UU^T = I$ and $VV^T = I$. $\Lambda$ is a diagonal matrix with its diagonal value as $\Lambda(i,i) = \lambda_i$. $J(P)$ is then transformed to:
\begin{equation}\small
J(P)= \mbox{tr}(2 P^T U\Lambda V^T ) = \mbox{tr}(2 V^T P^T U\Lambda  ).
\end{equation}
Define $Z =  V^T P^T U$. As $ZZ^T = I$, $Z$ is thus an orthogonal matrix. Therefore, $z_{ij}\leq 1$. As $J(P)$ can be rewritten as $J(P) =  \mbox{tr}(Z \Lambda ) = \sum_{i} z_{ii} \lambda_i \leq \sum_i \lambda_i$, the equivalence is achieved with $z_{ii} = 1$. Thus, when $Z$ is an identity matrix, the maximization is achieved. In this case, $Z = V^T P^T U = I$. Therefore the solution of the optimization problem defined in Eq. (\ref{Eq_Obj_Pc_Orth}) is 
\begin{equation}
P = UV^T.
\end{equation}
To this end, with the two closed form solutions, we can obtain the projection matrices to project the content features into the style features' subspace. This manifold alignment module can be plugged into many existing style transfer frameworks, which will be discussed in Section \ref{Sec_STF}.

\textbf{Extension to User Controlled Style Transfer.} 
The proposed algorithm can be easily extended to allow user editing or using segmentation maps as guidance. In both cases, the only change is to modify the affinity matrix $A^{cs}$. Denote the corresponding regions as $(\Omega_1^c, \Omega_1^s),..., (\Omega_i^c, \Omega_i^s),..., (\Omega_m^c, \Omega_m^s)$, where $m$ is the total number of corresponding regions. $\Omega_i^c$ and $\Omega_i^s$ are the $i$th corresponding regions indicated by users to have similar style patterns. We design two strategies to incorporate these guidance. In the first case, users' input, such as strokes indicating corresponding semantic parts, provides partial semantic meaning. We use another matrix $A^{cs'}$ which encodes dense connections between such user defined related regions.
\begin{equation}\label{Eq_A_Cross_UserEdit}\small
  A_{ij}^{cs'} = \left\{
\begin{aligned}
1  \quad & (i \in \Omega_1^{c}~~\text{and}~~j \in \Omega_1^{s})\; \mbox{or} \; ...( i \in \Omega_m^{c}~~\mbox{and}~~j \in \Omega_m^{s}) \\
0  \quad & \mbox{otherwise}\,. \\
\end{aligned}
\right.
\end{equation}
Then the affinity matrix is obtained with an `OR' operation performed between $A_{ij}^{cs}$ in Eq.~(\ref{Eq_A_Cross}) and $A_{ij}^{cs'}$. 
 
In the second case, if the content image and style image have large overlap in semantic segmentation regions, then we can also make use of segmentation maps as guidance. Different from the small areas drawn by users, semantic segmentation regions may cover large areas, 
\emph{i.e.}, 
the whole object in the image. If dense connection is used, one pixel may have many corresponding pixels in the affinity matrix, leading to blurry results. Therefore, instead, we calculate $k$ nearest neighbors within the same semantic region:
\begin{equation}\label{Eq_A_Cross_Semantic}\small
  A_{ij}^{cs} = \left\{
\begin{aligned}
1  \quad & \big( \phi_i(F_c) \in \mathcal{N}_k(\phi_j(F_s))~~\mbox{and}~~i \in \Omega_p^c, j \in \Omega_p^s \big)~~\mbox{or}\\
         & \big( \phi_j(F_s) \in \mathcal{N}_k(\phi_i(F_c))~~\mbox{and}~~i \in \Omega_p^c, j \in \Omega_p^s \big)  \\
0  \quad & \mbox{otherwise}\,.
\end{aligned}
\right. 
\end{equation}
With these different affinity matrix construction strategies, the following style transfer procedure is the same as before. 


\subsection{Style Transfer Framework}
\label{Sec_STF}
Currently, common style transfer frameworks include single-level and multi-level style transfer frameworks~\cite{li2017universal}, which are illustrated in Figures \ref{Fig_Framework_GST} (a) and (b). Although both frameworks produce promising results on artistic style transfer, they have limitation on photorealistic style transfer. This is because both frameworks are built on encoder-decoders, which will compress the content features and lead to the loss of content details. We therefore propose a new style transfer structure for photorealistic style transfer, which we call Adaptive Weight Skip Connection Structure (AWSC), shown in Figure \ref{Fig_Framework_GST} (c).

\textbf{Adaptive Weight Skip Connection Structure}. The main idea behind this structure is to align the low level content and style features produced by the encoder and skiply feed the aligned features to the decoder. As the low level features which have higher resolution encode more details of the image, the objective is to help decoder recover more details for photorealistic style transfer. Specifically, suppose the content and style features produced by the $i$th layer of the encoder are defined as $F_c^{(i)}$, $F_s^{(i)}$, and the aligned feature as $F_{cs}^{(i)}$. Denote the output of the decoder corresponding to the $i$th layer of the encoder as $\hat{F}_{cs}^{(i)}$, then the input to the $(i-1)$th adaptive weight skip connection module is $F_{out}^{i-1} = \hat{F}_{cs}^{(i)}$ and $F_{connect}^{i-1} = F_{cs}^{(i-1)}$. In the adaptive weight skip connection module, the mean and variance of $F_{out}^{i-1}$ will be used to adaptively normalize the features $F_{connect}^{i-1}$. After the operation, the normalized features and  $F_{out}^{i-1}$ are weighted with 0.5 and added as the input to the $(i-1)$th decoder layer. By introducing the AWSC module, the decoder will use both the features $F_{connect}$ that contain details and the features $F_{out}$ that contain more semantic information for producing photorealistic stylized results.

\section{Experimental Results}
In this section, ablation studies are carried out to demonstrate the effectiveness of our method. Qualitative and running time comparison with representative state-of-the-art methods and user study on both artistic and photorealistic style transfer are then presented.

\subsection{Experimental Setting}
For the encoder and decoder structures, we adopted the first 5 layers of the VGG-19 model~\cite{DBLP:journals/corr/SimonyanZ14a} pre-trained on ImageNet~\cite{deng2009imagenet} as the encoder. The decoder has the symmetric structure as the encoder. We trained the auto-encoder by reconstructing input images on the MS-COCO dataset~\cite{lin2014microsoft} and WikiArt dataset~\cite{duck2016painter}. For training the photorealistic style transfer network with the skip connection structure, MS-COCO dataset is used. While training the network with the introduced AWSC modules, one input training image is used as both content and style images. The network is trained with the reconstruction loss similar to an auto-encoder. For evaluation of artistic style transfer, images in the MS-COCO dataset \cite{lin2014microsoft} are used as content images and the WikiArt dataset \cite{duck2016painter} provides style images. For evaluation of photorealistic style transfer, images in the MS-COCO dataset \cite{lin2014microsoft} are used as both content and style. 

For the setting of $k$ nearest neighbors in Eq. (\ref{Eq_A_Cross}), we found larger $k$ typically leads to smooth or blurry results. However, no significant influence is observed when setting $k$ to a small number around 5. $k$ is thus set as 5. Another parameter is $\alpha$, which is the stylization weights $F_{cs} = (1-\alpha )F_{c} + \alpha F_{cs} $. It is set to 0.6 to balance between the content preservation and stylization level. 




\subsection{Ablation Study}

\begin{figure}[t]
  \centering
  \includegraphics[width=0.47\textwidth]{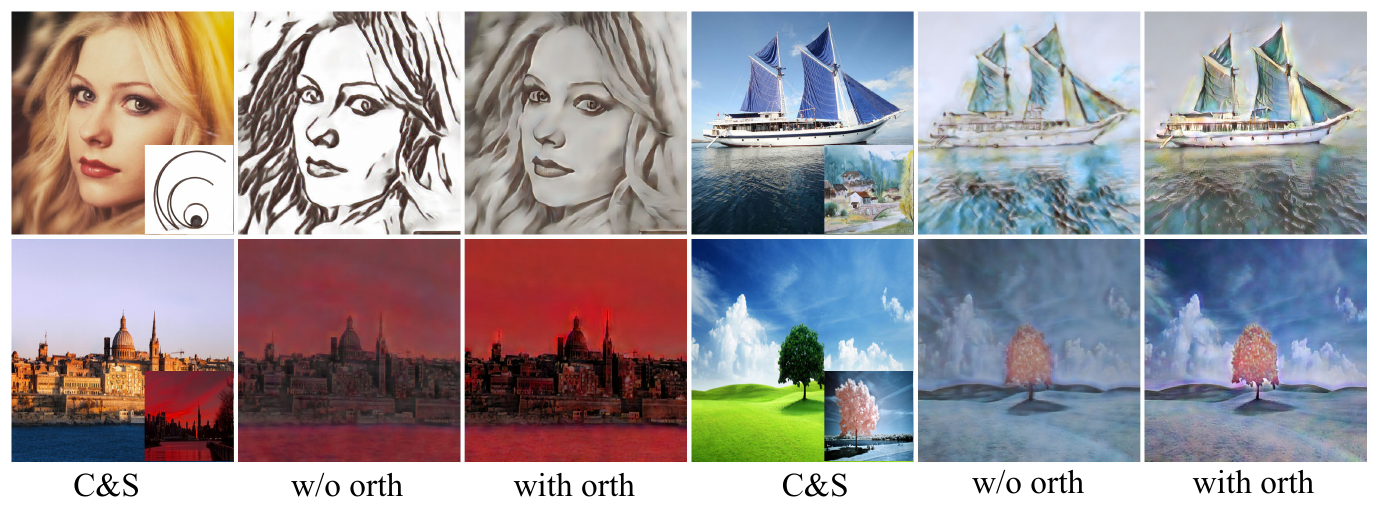}
 \setlength{\abovecaptionskip}{0.01cm}
  \caption{\small{The influence of orthogonal constraints on artistic style transfer (1st row) and photorealistic style transfer (2nd row).} }
  \label{Fig_Orth}
\end{figure}

\textbf{Influence of orthogonal constraints.} To study the influence of orthogonal constraints for both artistic and photorealistic style transfer, we have tested style transfer with and without the orthogonal constraints. The results are given in Figure \ref{Fig_Orth}. In the first row of Figure \ref{Fig_Orth} are the artistic style transfer results. We can see that the orthogonal constraints may lead to less stylized results for artistic style transfer. However, the content preservation is better. As artistic style transfer does not require full content preservation, we use the non-orthogonal version for the following artistic style transfer. For the photorealistic style transfer in the second row, it can be seen that with the orthogonal constraints, the transferred results are clearer, with better content preservation. Therefore, for photorealistic style transfer, we use the orthogonal constrained version of our method.

\begin{figure}[t]
  \centering
  \includegraphics[width=0.47\textwidth]{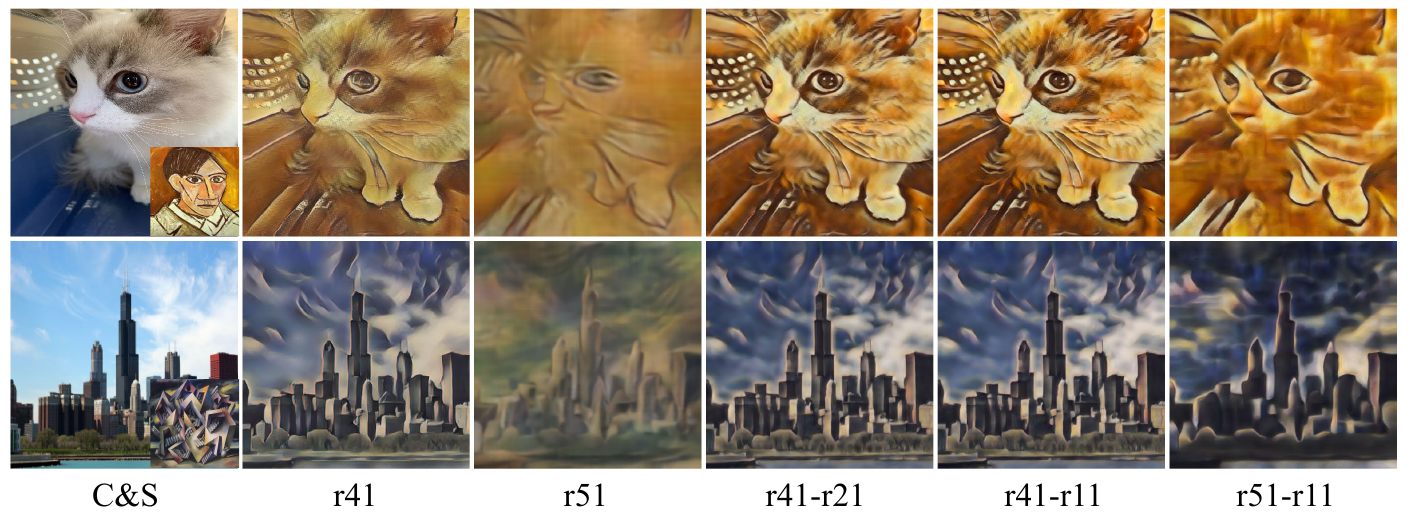}
    \setlength{\abovecaptionskip}{0.01cm}
  \caption{\small{Results of single-level and multi-level style transfer structures for artistic style transfer.} }
  \label{Fig_DiffLevel}
\end{figure}

\textbf{Influence of single level and multi-level style transfer structure on artistic style transfer.} To find the best style transfer structure for artistic style transfer, we have tested single level style transfer and multi-level style transfer structures as given in Figure \ref{Fig_Framework_GST}. Figure \ref{Fig_DiffLevel} shows the results, and the multi-level results are generally better than single level results. For single level style transfer, style transfer at a higher level, \textit{e.g.}, r51, may lead to blurry results. The results of multi-level style transfer of r51-r11 thus are also a little worse compared with the r41-r11 structure. However, the results of r41-r21 are almost the same as r41-r11. Meanwhile, r41-r21 is more efficient compared with r41-r11. Therefore, we use the multi-level style transfer at r41-r21 for artistic style transfer.

\begin{figure}[t]
  \centering
  \includegraphics[width=0.47\textwidth]{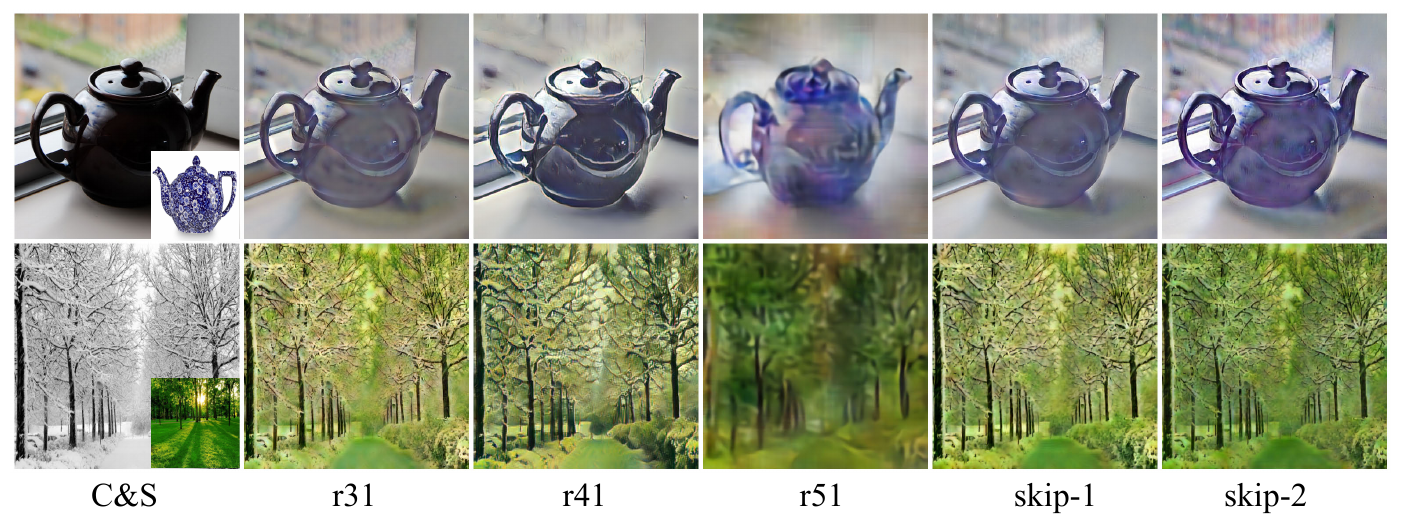}
    \setlength{\abovecaptionskip}{0.01cm}
  \caption{\small{The influence of the adaptive weight skip connection structure for photorealistic style transfer.} }
  \label{Fig_Skip}
\end{figure}



\begin{figure}[t]
  \centering
  \includegraphics[width=0.47\textwidth]{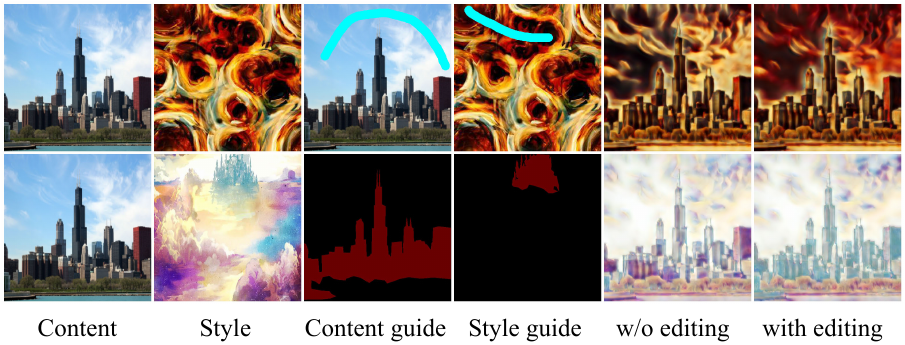}
    \setlength{\abovecaptionskip}{0.01cm}
  \caption{\small{User editing (the first row) and segmentation mask guidance (the second row) results. In the first and second columns are the input content and style images. The third and forth columns show the user drawn lines of corresponding regions or semantic segmentation masks. In the fifth and sixth columns are the original style transfer results and the results after using user guidance information.} }
  \label{Fig_Mask}
\end{figure}

\begin{figure*}[t]
  \centering
  \includegraphics[width=0.97\textwidth]{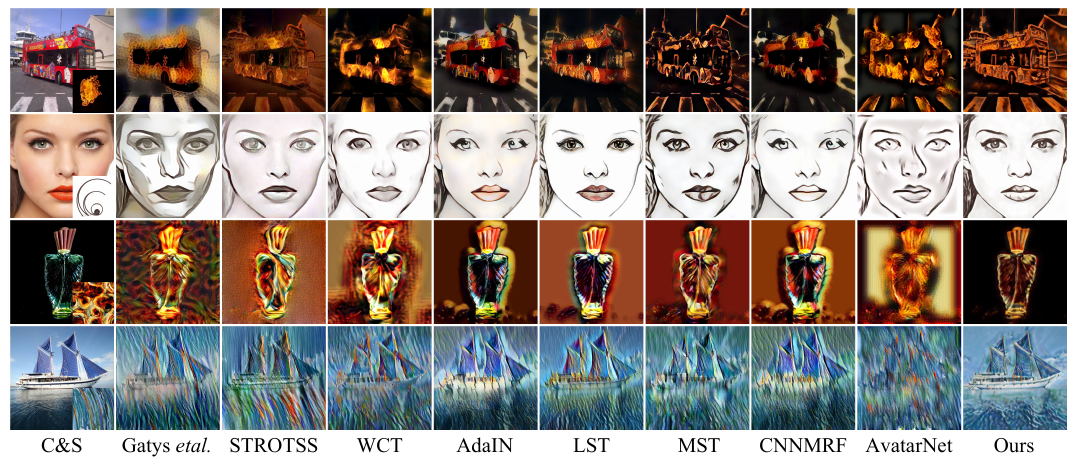}
    \setlength{\abovecaptionskip}{0.01cm}
  \caption{\small{Comparison with state-of-the-art methods for artistic style transfer. In the first column is the input content and style images (in the bottom right corner). The following columns are the corresponding stylized results of compared methods and the proposed method.} }
  \label{Fig_Comp_SOTA}
\end{figure*}

\textbf{Study of the adaptive weight skip connection structure.} We have tested two skip connection settings for photorealistic style transfer, with results given in Figure \ref{Fig_Skip}. One is to align the means of the two features $F_{out}$ and $F_{connect}$ (skip-1 in the figure) and the other is to align both the means and variances of the two features (skip-2 in the figure). For the skip connection, we have skip connected the r31-r41 layers. The best results of single level style transfer without skip connection is achieved at r41 layer. However, its result is worse than second skip connection setting (skip-2). This shows the effectiveness of the proposed skip connection mechanism for photorealistic style transfer.


\subsection{User Controlled Style Transfer Results}
In Figure \ref{Fig_Mask}, we show the user drawn editing results (the first row) and semantic segmentation guided editing (the second row) results. Comparing the fifth and sixth columns of Figure \ref{Fig_Mask} in the first row, we can see that with drawn corresponding areas on content and style images, the style transfer results have more desired appearance as specified by the user, \eg, the style of sky becomes more like the style defined by users on the style image. From the last row of Figure \ref{Fig_Mask}, results using segmentation maps as guidance are more semantically aligned compared with the original results. With the segmentation map introduced, although the buildings in the example style image occupy only a small area of the images, the proposed method can successfully transfer styles between the same semantic regions.


\subsection{Results of Artistic Style Transfer}

Qualitative comparisons with state-of-the-art methods for artistic style transfer are given in Figure~\ref{Fig_Comp_SOTA}. Compared methods include, the method of Gatys \etal \cite{gatys2016image}, STROTSS \cite{kolkin2019style}, WCT \cite{li2017universal}, AdaIN \cite{huang2017arbitrary}, LST \cite{li2019learning}, MST~\cite{Zhang_2019_ICCV}, CNNMRF~\cite{li2016combining} and AvatarNet~\cite{sheng2018avatar}. We have used the official codes of these methods, except for WCT, AdaIN and CNNMRF where their re-implementation versions are used. For all the compared methods, their original reported parameter settings are used.

Among the compared methods, Gatys \etal, WCT, AdaIN and LST are global statistics based methods. CNNMRF and AvatarNet are local patch based. STROTSS and MST are semantic based. For global statistics based methods, they fail to preserve the content structure when there is a large feature distribution difference between the style image and the content image, e.g., results in the first and last rows. Local patch based methods sometimes generate local artifacts due to local mismatches, which was also observed in some previous work~\cite{Zhang_2019_ICCV}. For semantic based methods, STROTSS sometimes mismatches style patterns to unwanted regions. For example, in the 2nd and the 4th examples, their results are obviously worse than ours. MST achieves relatively good results, but it requires the setting of cluster numbers. If improper cluster number is set, it produces degraded results. Besides, in some cases, it is hard to find clusters for style images, e.g., the 4th example. As can be seen, the proposed method which preserves the local feature affinity, produces apparently better stylized results.

\begin{table}[t]\small
\centering
\begin{tabular}{l l l l }
\toprule
Methods & CI & SL & Overall \\
\hline
Gatys & 6.773 & 6.311 & 6.286 \\
STROTSS & 8.303 & 7.076 & 7.765 \\
WCT & 6.731 & 7.176 & 6.756 \\
AdaIN & 6.748 & 6.555 & 6.622 \\
LST & 8.084 & 7.412 & 7.672 \\
MST & 7.420 & 6.916 & 6.983 \\
CNNMRF & 6.874 & 6.597 & 6.580 \\
Avatar-net & 5.580 & 6.639 & 5.891 \\
\hline
Ours & \textbf{8.479} & \textbf{7.815} & \textbf{8.176} \\
\bottomrule
\end{tabular}
    \setlength{\abovecaptionskip}{0.01cm}
\caption{\small{User study results of artistic style transfer, where `CI' represents content integrity and `SL' represents stylization level }}
\label{Table_Non-Photorealistic_user_study}
\end{table}

\begin{table*}[h]\small
\centering
\begin{tabular}{ccccccccccc}
\toprule
Methods   & Gatys \etal &STROTSS & WCT & WCT (r41-r21) & AdaIN & LST &MST & CNNMRF & Avatar-net & Ours \\
\hline
Time (s)  & 32.225 & 160.950 & 0.786  & 0.407 & \textbf{0.043}  & \textbf{0.040}  & 2.251 & 8.974 & 2.300 & \textbf{0.385} \\
\bottomrule
\end{tabular}
    \setlength{\abovecaptionskip}{0.01cm}
\caption{\small{Average running time (s) comparison with artistic style transfer methods.}}
\label{Table_Time}
\end{table*}

\begin{figure*}[h]
  \centering
  \includegraphics[width=0.97\textwidth]{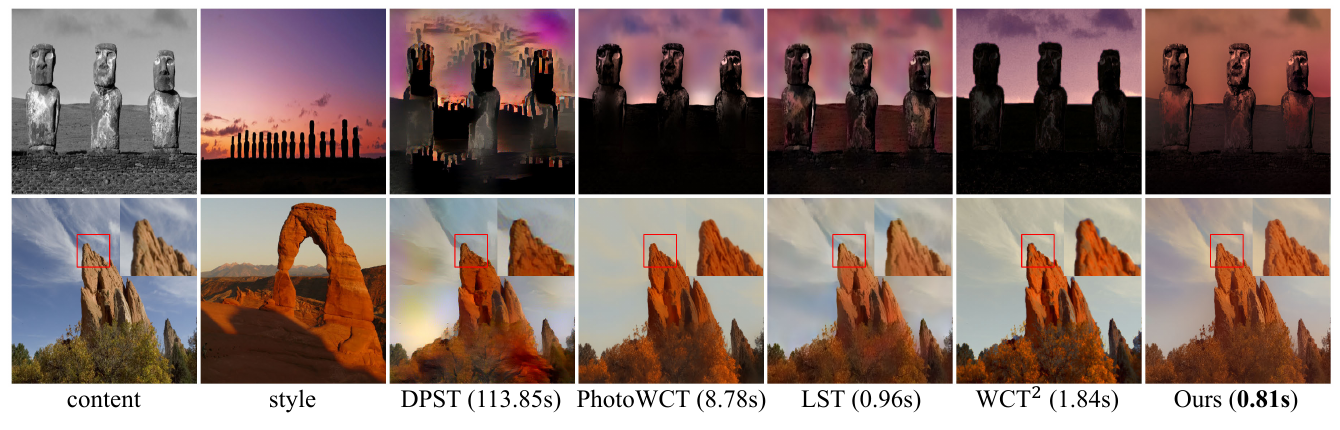}
    \setlength{\abovecaptionskip}{0.01cm}
  \caption{\small{Photorealistic style transfer results. In the 1st and 2nd columns are the content and style images respectively. The following columns are results of DPST, PhotoWCT, LST, WCT$^2$ and MAST, together with their average running time (s). Best view in large size.} }
  \label{Fig_Comp_SOTA_PR}
\end{figure*}

\textbf{User study.} For artistic style transfer, we used 25 content images and 32 style images to generate 800 images. We randomly selected 6 generated images for each method. The compared methods are the same with the qualitative comparison part. The results are presented to 20 users in a random manner. For each generation result, users were asked to rate the following three aspects, content integrity, stylization level and overall quality, with a score ranging from 1-10. For scoring content integrity, users were asked to measure if the content structure was corrupted. Stylization level is defined as the style similarity of the stylized image with the style image. The overall quality is defined as a combination of the above two evaluation metrics. With the collected responses, the average scores of each method are shown in Table~\ref{Table_Non-Photorealistic_user_study}. As can be seen, only STROTSS and LST among the compared methods are close to our method. The proposed method achieves the best results for all the three evaluation aspects. 

\textbf{Efficiency.} Running time comparison with artistic style transfer methods is reported in Table~\ref{Table_Time}. The results are the average running time of testing $10$ image pairs with the size of $512 \times 512$. All the methods are run on a server with an Intel Xeon CPU E5-2620 v4@2.10GHz and a GeForce RTX 2080 Ti GPU. Our algorithm uses the multi-level setting where style transfer is performed on r41-r21. Default settings are used for the compared methods, except for WCT which we have also tested its running time under the r41-r21 setting. We can see our algorithm is faster than most of the existing methods. It is also more efficient compared with the WCT under the same setting as our method. However, it is slightly less efficient than AdaIN and LST. There are two reasons for this phenomenon. One is that the two methods are single level based methods. The second is that our method requires the computation of affinity matrix, which is a bit time consuming. But our running time should be sufficient for practical applications.

\begin{table}[t]\small
\centering
\begin{tabular}{l l l l }
\toprule
Methods & PRL & SL & Overall\\
\hline
DPST & 6.698 & 6.920 & 6.653 \\
PhotoWCT & 7.809 & \textbf{7.548} & 7.548 \\
LST & 7.593 & 7.111 & 7.181 \\
WCT$^2$ & 7.236 & 7.322 & 7.166 \\
\hline
Ours & \textbf{8.181} & 7.452 & \textbf{8.075} \\
\bottomrule
\end{tabular}
    \setlength{\abovecaptionskip}{0.01cm}
\caption{\small{User study results of photorealistic style transfer, where `PRL' represents photorealistic level and `SL' represents stylization level.}}
\label{Table_Photorealistic_user_study}
\end{table}
\vspace{0.1cm}
\setlength{\belowcaptionskip}{0.1cm}


\subsection{Results of Photorealistic Style Transfer.} Qualitative comparisons with state-of-the-art methods for photorealistic style transfer are given in Figure~\ref{Fig_Comp_SOTA_PR}, along with their running time. The compared methods include DPST~\cite{luan2017deep}, PhotoWCT~\cite{li2018closed}, LST~\cite{li2019learning} and WCT$^2$~\cite{yoo2019photorealistic}. Except for DPST, we have used the official code, following their original settings. For DPST, the re-implemented version is used, also following its original setting. For the proposed method, we used the adaptive weight skip connection setting skip connecting r31 and r41. Both mean and variance are aligned before skip connection. The same post processing step proposed in PhotoWCT is adopted in our method. For all the compared methods, the results are generated with semantic segmentation maps considered.


As can be seen from the results, the DPST method may sometimes destroy the original content structure. The results of LST, WCT$^2$, PhotoWCT and ours are in fact similar. However, from the enlarged patch, the result of LST is a little blurry and there are usually some weird artifacts of WCT$^2$ occurring at the edges of objects. Overall, the proposed method has an advantage over the others in terms of semantic alignment, \textit{e.g.}, the sky in both examples look more like the style images. As for the running time comparison, our method is the fastest compared with the others. Only LST is comparable with our method. 


\textbf{User study.} We randomly selected 10 stylized results for each method, and presented them to 20 users in a random manner. For each combination, the users were asked to rate the following three metrics using a score ranging from 1-10, photorealistic level, stylization level and overall quality. The definition of photorealistic level is somewhat different from the content integrity metric used for artistic style transfer. For content integrity, small distortion without destroying the content structure in the stylized image is allowed. However, for photorealistic level, such distortion is not allowed. The average results are shown in Table~\ref{Table_Photorealistic_user_study}. From the table, the proposed MAST method is the best for photorealistic level and overall quality. For the stylization level, it is slightly worse compared with PhotoWCT, with the score of our method (7.452) and the score of PhotoWCT (7.548).

\section{Conclusion}
A new assumption that style transfer can be defined as two manifold distributions' alignment problem is made. Furthermore, a manifold alignment based style transfer (MAST) framework is proposed. The framework is shown to be flexible to allow for user controlled style transfer. Besides, the framework is further extended with a new adaptive weight skip connection structure for photorealistic style transfer. Extensive experiments verify the proposed method is efficient and achieves promising results compared with previous state-of-the-art methods on both artistic and photorealistic style transfer. 

\noindent\textbf{Acknowledgments.} This work is supported by Science and Technology Innovation 2030 New Generation Artificial Intelligence Major Project (2018AAA0100905), the National Natural Science Foundation of China (61806092), Natural Science Foundation of Jiangsu Province (BK20180326), the Fundamental Research Funds for the Central Universities (02021438008) and the Collaborative Innovation Center of Novel Software Technology and Industrialization.


{\small
\bibliographystyle{ieee_fullname}
\bibliography{egbib}
}

\end{document}